\documentclass[letterpaper]{article} 
\usepackage{aaai24}  
\usepackage{times}  
\usepackage{helvet}  
\usepackage{courier}  
\usepackage[hyphens]{url}  
\usepackage{graphicx} 
\usepackage{natbib}  
\usepackage{caption} 
\usepackage{listings}
\DeclareCaptionStyle{ruled}{labelfont=normalfont,labelsep=colon,strut=off} 
\usepackage{times}
\usepackage{epsfig}
\usepackage{amsmath}
\usepackage{amssymb}
\usepackage{multirow}
\usepackage{makecell}
\usepackage{mathtools}
\usepackage{kotex}
\usepackage{algorithm}
\usepackage[noend]{algpseudocode}
\usepackage[dvipsnames]{xcolor}
\usepackage{mathabx}

\title{MetaMix: Meta-state Precision Searcher for Mixed-precision Activation Quantization}
\author{
    Han-Byul Kim\textsuperscript{\rm 1,2}\thanks{Work done during internship at Google},\; 
    Joo Hyung Lee\textsuperscript{\rm 2},\; Sungjoo Yoo\textsuperscript{\rm 1},\; Hong-Seok Kim\textsuperscript{\rm 2}
}
\affiliations{
    \textsuperscript{\rm 1}Department of Computer Science and Engineering, Seoul National University, Seoul, Korea \\
    \textsuperscript{\rm 2}Google, Mountain View, California, USA \\
    shinestarhb@gmail.com, rinolee@google.com, sungjoo.yoo@gmail.com, hongseok@google.com
}

\begin{document}

\maketitle

\begin{abstract}
Mixed-precision quantization of efficient networks often suffer from activation instability encountered in the exploration of bit selections. To address this problem, we propose a novel method called MetaMix which consists of bit selection and weight training phases. The bit selection phase iterates two steps, (1) the mixed-precision-aware weight update, and (2) the bit-search training with the fixed mixed-precision-aware weights, both of which combined reduce activation instability in mixed-precision quantization and contribute to fast and high-quality bit selection. The weight training phase exploits the weights and step sizes trained in the bit selection phase and fine-tunes them thereby offering fast training. Our experiments with efficient and hard-to-quantize networks, i.e., MobileNet v2 and v3, and ResNet-18 on ImageNet show that our proposed method pushes the boundary of mixed-precision quantization, in terms of accuracy vs. operations, by outperforming both mixed- and single-precision SOTA methods.
\end{abstract}

\section{Introduction}

The ever increasing demand of efficiency requires the quantization of efficient models, e.g., MobileNet-v1~\cite{mbv1}, v2~\cite{mbv2} and v3~\cite{mbv3}, in lower precision. 
Mixed-precision quantization looks promising due to the supports available on existing computing systems, e.g., 8-bit, 4-bit and 1-bit integer~\cite{nvturing,bitfusion}.

In this work, we address per-layer mixed-precision quantization for activation while utilizing a single bit-width of weight.
Mixed-precision quantization of efficient models, however, is challenging in that the design space of bit-width selection (in short, bit selection) is prohibitively large. Especially, each candidate of bit selection needs to be trained, e.g., on ImageNet dataset, for evaluation, which incurs prohibitively high training cost.

In order to address the training cost problem, we often explore bit selections while training the network weights~\cite{hmq,dq,djpq,haq,fracbits,sdq,nipq,dnas,edmips,bpnas}. 
In such a combination of bit selection and weight training, the bit-width of each layer can be changed (due to bit selection) across training iterations. Such a bit-width change during model training incurs a new problem called {\it activation instability due to bit selection}.

\begin{figure*}
\begin{center}
\includegraphics[width=0.9\linewidth]{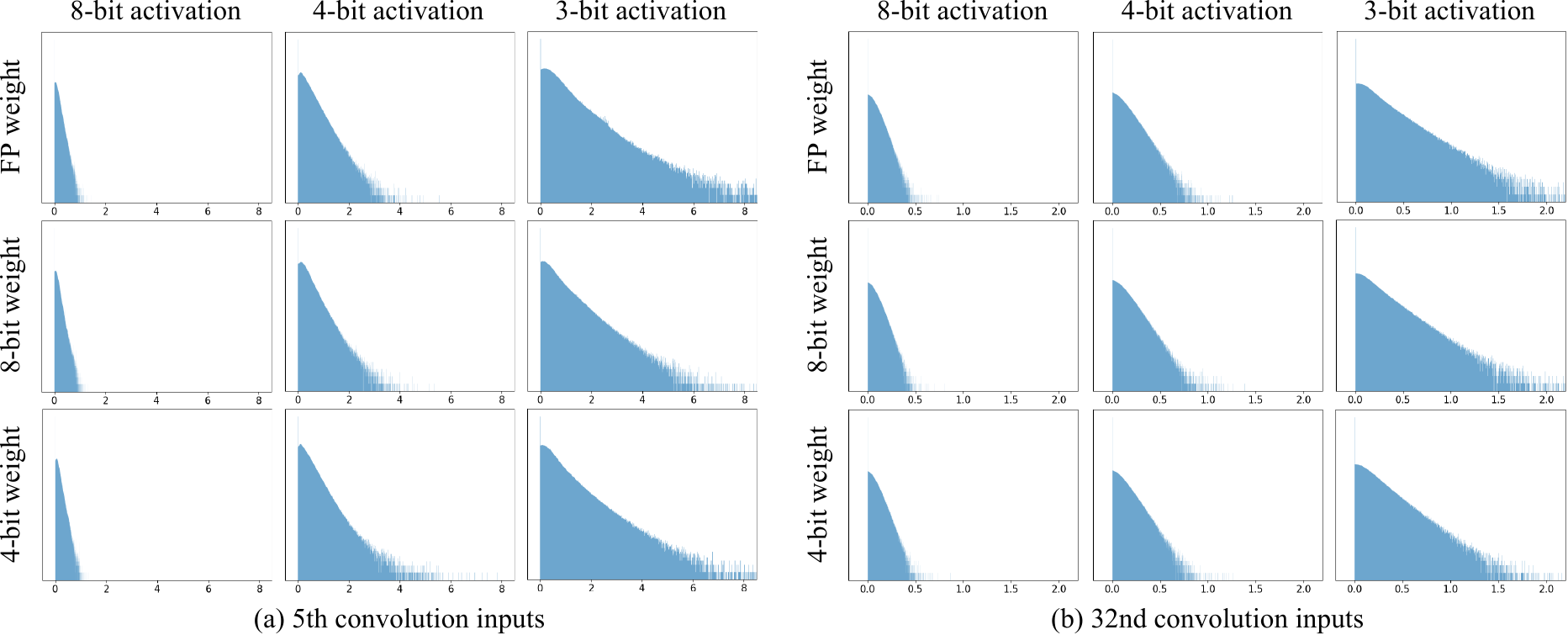}
\end{center}
  \caption{
  Input activation distribution before quantizer with 8-bit, 4-bit, and 3-bit single-precision activation quantization of MobileNet-v2
  (a) in 5th layer (depth-wise convolution in 2nd block) and 
  (b) in 32nd layer (depth-wise convolution in 11th block).
  Each row has the same fixed weight bit-width and each column has the same fixed activation bit-width.
  `FP' represents full-precision.
  In all the figures, x-axis is for values and y-axis is for frequency in log scale.
  }
\label{fig:fixedbit_actdist}
\end{figure*}

Figure \ref{fig:fixedbit_actdist} illustrates the input activation distribution before quantizer in the MobileNet-v2 model across different bit-widths of weight and activation. 
The figure shows, for a given activation bit-width, the activation distribution remains consistent across different weight bit-widths, e.g., the distribution of 8-bit activation across FP (full-precision), 8-bit and 4-bit weights in the first columns.
However, given a bit-width (e.g., 8-bit) of weight, the distribution of the activation significantly varies across different bit-widths (e.g., 8-bit, 4-bit and 3-bit) of activations, which demonstrates activation instability due to bit selection.

Activation instability, due to weight quantization, has been reported in previous works~\cite{profit,fqnet}. The problem is encountered when single-precision models are quantized in low bits. 
In this paper, we report a new activation instability problem encountered in the exploration of bit selections under model training. In order to address these activation instabilities, we propose a new training method called {\it MetaMix}.
Our contribution is summarized as follows.

{\begin{itemize}
    \item We demonstrate a new problem called {\it activation instability due to bit selection} which disrupts exploring bit selection while training the network thereby offering sub-optimal results.
    \item Our proposed MetaMix mitigates activation instability problem. MetaMix consists of two phases: bit selection and weight training. The bit selection phase offers fast and high-quality bit selection by both bit-meta training step and bit-search training step.
    \item In the bit-meta training step, we train the network weights at multiple bit-widths to obtain a state called {\it meta-state} which provides consistent activation distribution across different activation bit-widths. Then, in the bit-search training step, on the fixed meta-state, we learn architectural parameters for per-layer bit-width probabilities. We iterate both steps, which proves effective for fast and high-quality bit selection while mitigating the effect of activation instability. 
    \item In the weight training phase, we continue to train, i.e., fine-tune network weights and step sizes\footnote{The size of quantization range is determined by $(2^{b}-1)*s$ in case of $b$-bits and a step size of $s$.} under the fixed per-layer bit-widths previously determined in the bit selection phase, which offers fast training due to the mixed-precision-aware initialization of weights and step sizes done in the previous phase.
    \item We evaluate our method on highly optimized and hard-to-quantize networks, i.e., MobileNet-v2 and v3 and ResNet-18 on ImageNet-1K dataset~\cite{imgnet} and show that ours offers state-of-the-art results.
\end{itemize}}

\section{Related Works}

Uniform quantization~\cite{dorefa,pact,lsq} is hardware friendly since most of compute devices, e.g., GPU, support uniform grids (or levels).
Non-uniform quantization~\cite{lqnet,apot,lcq,n2uq} optimizes quantization grids in order to fit diverse distributions thereby enabling lower bit-widths while requiring specialized compute devices.
In our work, we mainly target uniform method.

{ \bf Trainable quantization: }
Training quantization parameters~\cite{pact,qil,lsq} like clip range or step size has a potential of lower precision and thus its possibilities have been actively investigated.
Most of these works show good results in relatively redundant networks, e.g. ResNets, but fail to quantize highly optimized networks, e.g. MobileNet-v2 and v3, in 4-bit without accuracy loss.
Recently, PROFIT~\cite{profit} offers 4-bit quantization in MobileNets with a special training recipe to address activation instability due to weight quantization. 
BASQ~\cite{basq} enables low precision for MobileNets via quantization hyper-parameter search.

{ \bf Mixed-precision quantization: }
Existing mixed-precision methods can be classified into four categories.
① Learning-based solutions~\cite{dq,djpq,fracbits,ddq} optimize bit-widths as learnable parameters trained with gradient from task loss.
SDQ~\cite{sdq} adopts differentiable parameter for bit-width probability.
NIPQ~\cite{nipq} extends DiffQ~\cite{diffq} to mixed precision.
② RL-based solutions~\cite{haq,releq} train an RL agent which learns bit-width assignment policy.
③ NAS-based solutions~\cite{dnas,spos,bpnas,edmips} explore the design space of bit selection and attempt to solve selection problems via evolutionary search, differentiation, etc.~\cite{enas,darts,proxylessnas,dsnas}.
④ Metric-based solutions determine bit-width based on statistics, i.e., Hessian spectrum~\cite{hawqv1,hawqv2,hawqv3}.

In this paper, we present a novel NAS-based mixed-precision method. Ours tries to overcome the limitations of existing NAS-based solutions, e.g., high search cost and sub-optimal results due to activation instability.

\section{Activation Instability on Mixed-Precision Quantization}
\label{section:activation_instability}

\begin{figure}[t]
    \begin{center}
        \includegraphics[width=1.0\linewidth]{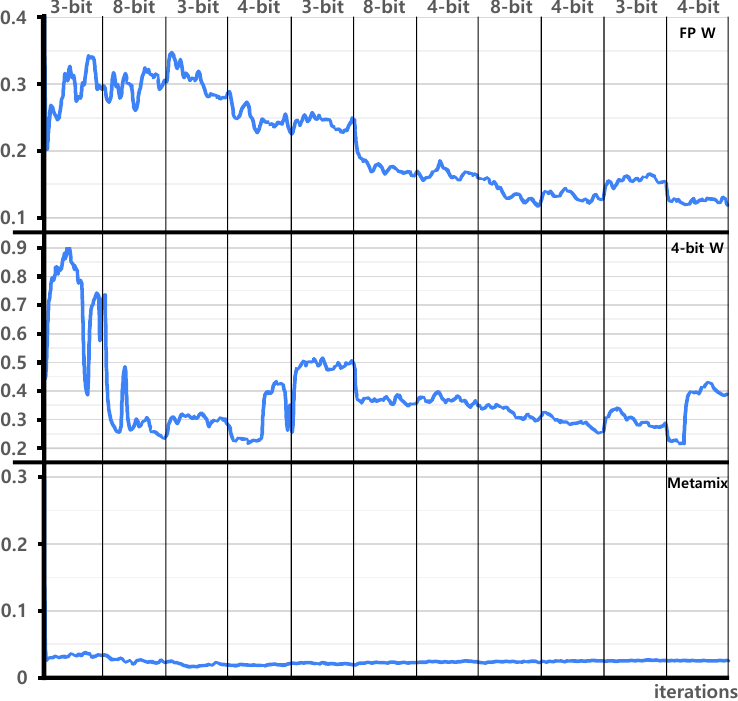}
    \end{center}
    \caption{Trend of batch norm statistics over iterations when changing activation bit-width. We plot the running variance of batch norm which follows 5th (depth-wise) convolution layer in 2nd block (Top: FP weights, Middle: 4-bit weights, Bottom: applying MetaMix with FP weights).}
    \label{fig:activation_stats}
\end{figure}

As Figure \ref{fig:fixedbit_actdist} illustrates, different activation bit-widths can yield different distributions even after normalization.
Figure \ref{fig:activation_stats} exemplifies how the activation distributions vary in the process of bit selection under weight training.
We change the activation bit-width of one layer of MobileNet-v2 every epoch of training by randomly picking the bit-width within 8-bit, 4-bit, and 3-bit.
We experiment two cases of weight bit-width: full-precision (top of Figure \ref{fig:activation_stats}) and 4-bit (middle).
The figure shows the running variance of batch normalization layer which follows the quantized depth-wise convolution layer inside of inverted residual block.

Figure \ref{fig:activation_stats} shows two trends in terms of activation instability.
First, activation statistics has strong correlation with bit-width.
Second, the correlation can be amplified when both activation and weight are trained and quantized in low bits.
As the figure demonstrates, in mixed-precision quantization, activation instability results from two sources, one from bit selection and the other from weight quantization.

Activation instability due to weight quantization results from the fact that the output activations of a layer can become different, even when the same input activations are used, due to weight update and subsequent weight quantization~\cite{profit}. Activation instability due to bit selection results from the fact that the lower precision tends to incur the more variance in the quantized data. For details about the instability, refer to the supplementary\footnote{Refer to our arXiv version for the supplementary materials.}.

As will be shown in our experiments, the activation instability can yield poor quality of mixed-precision network.
It is because, when the statistics of batch norm layer significantly vary due to activation instability during training, it is challenging to obtain a representative activation statistics\footnote{Note that the activation statistics obtained in training time is used in the batch norm layer in test time.} 
on the candidate bit-widths (during bit selection) as well as the final bit-widths (selected as a result of bit selection).
In \cite{profit,fqnet}, in order to cope with activation instability due to weight quantization, the authors propose fine-tuning the quantized network (with a single precision) to stabilize batch norm statistics at the end of training. However, according to our experiments (also to be mentioned later), it does not prove effective in mixed-precision quantization. It is mainly because sensitive layers like depth-wise convolution tend to be assigned high precision, which makes activation instability due to weight quantization less significant on those layers thereby reducing the effects of fine-tuning.
In this paper, we propose MetaMix to tackle the activation instability in mixed-precision quantization.
MetaMix effectively stabilizes batch norm statistics (bottom of Figure \ref{fig:activation_stats}) thereby providing representative activation statistics (Figure \ref{fig:metamix_actdist}) and high-quality bit selection (Figure \ref{fig:bits_mbv2}).

\section{MetaMix – a Meta-State Precision Searcher}

\subsection{Overall Training Flow}
\label{section:overall_flow}

Figure \ref{fig:metamix_flow_diagram} shows the overall training flow. 
Given a trained network, our proposed MetaMix determines per-layer bit-width of activations in the bit selection phase and fine-tunes network weights and step sizes in the weight training phase.

Algorithm \ref{alg:metamix} shows the overall process of bit selection phase.
The bit selection phase iteratively executes two steps: {\it bit-meta training} and {\it bit-search training}.
The bit-meta training step trains network weights in a mixed-precision-aware manner.
The bit-search training step learns the architectural parameters for per-layer bit-width probabilities on the fixed mixed-precision-aware weights learned in the bit-meta training step.
In the weight training phase, using the per-layer bit-widths determined in the bit selection phase, we fine-tune the weights and step sizes for both weights and activations.

Figure \ref{fig:metamix_flow_diagram} also shows that, given a network, we augment each layer with multiple branches, each for a specific bit-width, for the purpose of bit selection.\footnote{In the weight training phase, the original network architecture is utilized since only one of the branches is selected in the bit selection phase.}
Figure \ref{fig:searchstrategy_block} illustrates the details of multi-branch block. 
We augment the quantizer into $B$ branches where $B$ is the number of available bit-widths.
In the figure, we assume three bit-widths and thus obtain three branches for 8-bit, 4-bit and 3-bit.
Each branch is also associated with an architectural parameter for the bit-width probability.
Unlike the multi-branch quantizers, we do not duplicate but share weights across the branches.

\subsection{Bit-Meta Training}
\label{section:bit-meta}

\begin{figure}[t]
    \begin{center}
        \includegraphics[width=\linewidth]{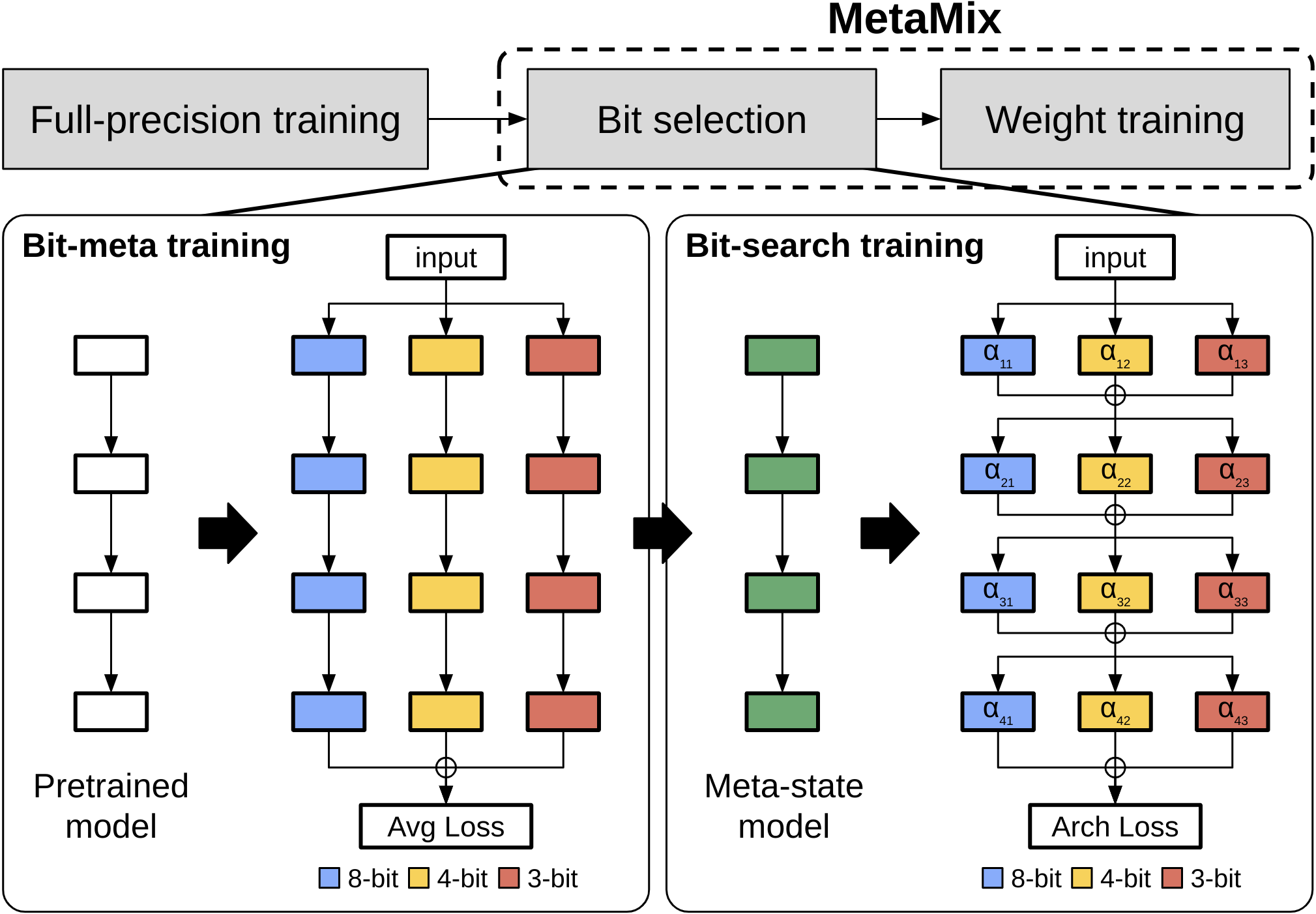}
    \end{center}
    \caption{MetaMix flow diagram and working mechanism.}
    \label{fig:metamix_flow_diagram}
\end{figure}

\begin{figure}[t]
    \begin{center}
        \includegraphics[width=\linewidth]{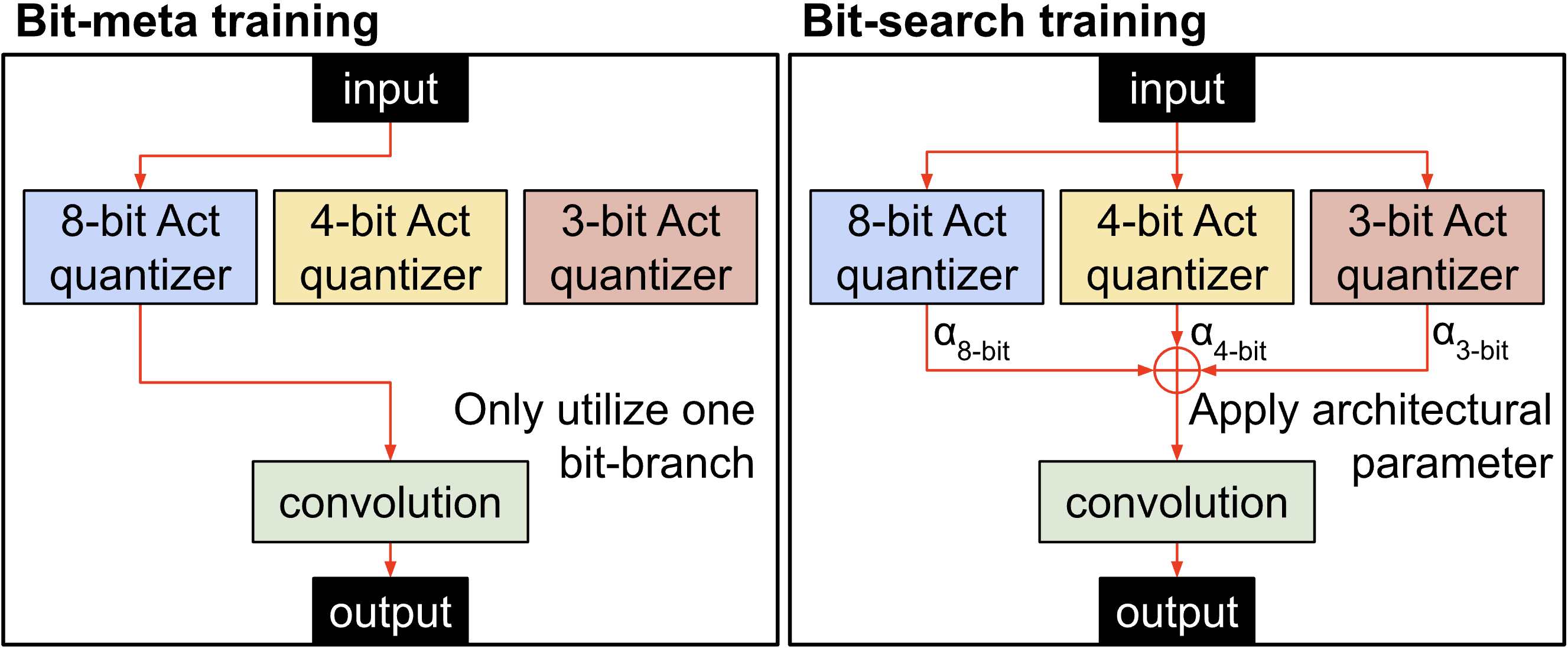}
    \end{center}
    \caption{MetaMix block structure design and operations on bit selection phase. ‘Act’ represents activation.}
    \label{fig:searchstrategy_block}
\end{figure}

Bit-meta training step learns network weights to be used in the subsequent bit-search training.
In order for bit-search to be successful, the network weights need to be learned in a mixed-precision-aware manner in order to facilitate the training of architectural parameters for per-layer bit-width probabilities in the bit-search training step.
Since our problem of learning network weights in the bit-meta training step is similar to that of meta-learning, e.g., MAML~\cite{maml}, which tries to facilitate fine-tuning for the given target task (analogous to the training of architectural parameters for bit-width probabilities in the bit selection phase in our case),
we apply meta-learning to learn our mixed-precision-aware weights and thus this step is called bit-meta training.

Equation \ref{eq:bitmeta} shows the loss of bit-meta training.
We select a bit-width $b_i$ among $B$ candidate bit-widths in end-to-end network and utilize it to build a single-precision network during the forward pass of training. After obtaining all the training losses ($B$ losses from $B$ candidate bit-widths), 
we utilize the total average loss in Equation \ref{eq:bitmeta} to update the network weight $w$.

\begin{equation}
\label{eq:bitmeta}
\mathcal{L}(w) = {1 \over B} \sum_{i = 1}^{B}{\mathcal{L}(w, b_i)}
\end{equation}
\vspace{+0.1cm}

We call the mixed-precision-aware weights obtained from the bit-meta training {\it meta-state} and the model with the meta-state {\it meta-state model} as illustrated in Figure \ref{fig:metamix_flow_diagram}.
Our key idea behind the bit-meta training is to minimizing the negative impact of activation bit-width to weight on bit selection phase as much as possible.
The mitigation is realized by training the weights in the bit-meta training and fixing them during the bit-search training. 
Thus, unlike existing methods (all the methods in Figure \ref{fig:bops}) which suffer from activation instabilities by exploring bit selections under the weights affected by bit-width, our proposed method can mitigate activation instabilities, by keeping model to search bit selection on meta-state model, thereby outperforming the existing methods as will be shown in the experiments.

\begin{algorithm}[H]
    \caption{Pseudo code of MetaMix (bit selection phase)}
    \label{alg:metamix}
    \textbf{Input}: the number of training iterations in 1-epoch $\mathcal{T}$, batched input image $X$, activation bit-width selection candidates $\{b_1,\cdots,b_B\}$ and the number of candidates $B$, per-bit \& per-layer architectural parameter $\alpha$, per-bit \& per-layer activation quantizer $q()$, network weight $w$, number of layers in model $L$, L1 regularization term $\textit{r}()$ and regularization scale factor $\lambda_r$ \\
    \textbf{Output}: trained weights, determined per-layer bit selections
    \hrule
    \vspace{+0.05cm}
    \begin{algorithmic}[1]
        \Function{b-meta-fwd}{$\textit{w}, input, \textit{b}_i, i$} ${\color{RoyalBlue} \; \cdots \; \textnormal{Fig.3(left)}}$
            \State $x_0 = \textnormal{input}$
            \For{\texttt{$l = 1 : L$}}
                \State $x_l = \textit{Forward}(w_l, q_{l,i}(x_{l-1}, b_i)) {\color{RoyalBlue} \; \cdots \; \textnormal{Fig.4(left)}}$
            \EndFor
            \Return $x_L$
        \EndFunction
    \end{algorithmic}
    \vspace{+0.05cm}
    \hrule
    \vspace{+0.05cm}
    \begin{algorithmic}[1]
        \Function{b-search-fwd}{$\textit{w}, input, \{\textit{b}_1, \cdots, \textit{b}_B\}, \; \alpha$}
            \State $x_0 = \textnormal{input} {\color{RoyalBlue} \; \cdots \; \textnormal{Fig.3(right)}}$
            \For{\texttt{$l = 1 : L$}} $\quad\quad\quad\quad\quad\quad\quad\quad\quad {\color{RoyalBlue} \ursh \quad Eq.2}$
                \State $\bar{q_l}(x_{l-1}) = \sum_{i = 1}^{B}{ \exp(\alpha_{l,b_i}) \over \sum_{j = 1}^{B}{\exp(\alpha_{l,b_j})}} \cdot q_{l,i}(x_{l-1}, b_i)$
                \State $x_{l} = \textit{Forward}(w_l, \bar{q_l}(x_{l-1})) {\color{RoyalBlue} \; \cdots \; \textnormal{Fig.4(right)}}$
            \EndFor
            \State \Return $x_L$
        \EndFunction
    \end{algorithmic}
    \vspace{+0.05cm}
    \hrule
    \vspace{+0.05cm}
    \begin{algorithmic}[1]
        \State {\color{OliveGreen} \textbf{\textit{\# Bit selection phase}}}
        \For{\texttt{$epoch = 1 : 2$}}
            \For{\texttt{$iter = 1 : \mathcal{T}$}}
                \State {\color{OliveGreen} \textbf{\textit{\# Bit-meta training}}}
                \For{\texttt{$i = 1 : B$}}
                    \State $out = \textnormal{\small B-META-FWD}\ (w, X_{iter}, b_i, i)$ 
                    \State $\mathcal{L}(w, b_i) = Loss(out)$
                \EndFor
                \State $\mathcal{L}(w) = {1 \over B} \sum_{i=1}^{B}\mathcal{L}(w, b_i) \quad {\color{RoyalBlue}\cdots \quad Eq.1}$
                \vspace{+0.1cm}
                \State $w \leftarrow \textit{Backward}(\mathcal{L}(w))$ \quad {\color{Gray} \textbf{\textit{\# update weight}}}
                
                \State {\color{OliveGreen} \textbf{\textit{\# Bit-search training}}}
                \If{\texttt{$epoch > 1$}}
                    \State $out = \textnormal{\small B-SEARCH-FWD} (w, X_{iter}, \{b_1, \cdot\cdot, b_B\}, \alpha)$ 
                    \State $\mathcal{L}(w,\alpha) = Loss(out)$
                    \State $\mathcal{L}(\alpha) = \mathcal{L}(w,\alpha) + \lambda_r \cdot \textit{r}(\alpha) {\color{RoyalBlue} \quad \cdots \quad Eq.3}$
                    \State $\alpha \leftarrow \textit{Backward}(\mathcal{L}(\alpha))$ \quad {\color{Gray} \textbf{\textit{\# update architectural param}}}
                \EndIf
            \EndFor
        \EndFor
        \For{\texttt{$l = 1 : L$}}
            \State $\textit{bit\_sel}_l = \textnormal{argmax}_b \alpha_l$ \quad {\color{Gray} \textbf{\textit{\# max $\alpha$ as per-layer bit}}}
        \EndFor
        \State \textbf{return} $W, \textit{bit\_sel}$ \quad {\color{Gray} \textbf{\textit{\# pass to next phase}} }
    \end{algorithmic}
\end{algorithm}

Note that both the iteration of bit-meta and bit-search training and the usage of fixed weights in bit-search training contribute to the reduction of activation instabilities during bit selection.
Specifically, the bit-meta training can reduce activation instability due to bit selection (Figure \ref{fig:metamix_actdist}), which benefits the subsequent bit-search training.
In addition, the usage of fixed full-precision weights in the bit-search training allows the bit-search training to avoid activation instability due to both bit selection and weight quantization, which contributes to the quality of selected bit-widths (Figure \ref{fig:bits_mbv2}).

\begin{table*}
    \begin{center}
        \begin{tabular}{|c|c|c|c|}
        \hline
        Phase & Epochs & Step / Network & Train or Fix? \\
        \hline\hline
        
        \multirow{3}{*}{\makecell{Bit\\selection}} & epoch \#1 & Bit-meta training & \makecell[l]{Train FP weight and $s_a$ with fixed $\alpha$} \\
        \cline{2-4}
        \multirow{2}{*}{} & epoch \#2 & \makecell{Bit-meta training $\rightleftarrows$ Bit-search training\\(two steps alternate for 1 iteration each)} & \makecell[l]{Bit-meta: Train FP weight and $s_a$ with fixed $\alpha$ \\ Bit-search: Train $\alpha$ with fixed FP weight and $s_a$} \\
        \hline
    
        \multirow{2}{*}{\makecell{Weight\\training}} & epoch \#3 - \#140 & MobileNet-v2 and v3 & \makecell[l]{Fine-tuning 4-bit weight, $s_a$ and $s_w$}\\
        \cline{2-3}
        \multirow{2}{*}{} & epoch \#3 - \#90 & ResNet-18 & \makecell[l]{Fixed per-layer activation bit-width (max $\alpha$)} \\
    
        \hline
        \end{tabular}
    \end{center}
    \caption{Detailed training process
    ($\alpha$: architectural parameters, $s_a$: step sizes of per-layer activations, $s_w$: step sizes of weights)}
    \label{table:training_details}
\end{table*}

\subsection{Bit-Search Training}
\label{section:bit-search}

Bit-search training learns an architectural parameter for per-layer bit-width probability on each of the branches in the block structure of Figure \ref{fig:searchstrategy_block}. Given an activation $x$, its quantized activation $\bar{q}(x)$ is calculated as follows.

\begin{equation}
\label{eq:bitsearch_darts}
    \bar{q}(x) = \sum_{i = 1}^{B}{ \exp(\alpha_{b_i}) \over \sum_{j = 1}^{B}{\exp(\alpha_{b_j})}} \cdot q_i(x, b_i)
\end{equation}
\vspace{+0.1cm}

Each branch with $b_i$-bits is assigned an architectural parameter $\alpha_{b_i}$. We first obtain the branch-specific quantized activation, $q_i(x, b_i)$ and then calculate $\bar{q}(x)$ by weighting the branch activation with its softmax as shown in Equation \ref{eq:bitsearch_darts}.

Note that the branches are differently utilized on bit-meta and bit-search training.
As Figures \ref{fig:metamix_flow_diagram} and \ref{fig:searchstrategy_block} (left) show, in bit-meta training, for a bit-width, only the associated branch is utilized without architectural parameter $\alpha_{b_i}$ to build a single-precision network where all the layers have the same bit-width.
However, in bit-search training, as Equation \ref{eq:bitsearch_darts} and Figures \ref{fig:metamix_flow_diagram} and \ref{fig:searchstrategy_block} (right) show, each branch obtains its own quantized result $q_i(x, b_i)$ on the given input $x$.
All the branches are utilized by weighting softmax of assigned $\alpha_{b_i}$.

Note also that, in this step, we utilize the model with fixed full-precision weights, i.e., the fixed meta-state model.
Thus, during back-propagation, we update only the architectural parameters to minimize the training loss
without updating the network weights. 
Equation \ref{eq:bitsearch_obj} shows the training loss.

\begin{equation}
\label{eq:bitsearch_obj}
    \mathcal{L}(\alpha) = \mathcal{L}(w, \alpha) + \lambda_{r} \cdot \textit{r}(\alpha)
\end{equation}
\vspace{+0.01cm}

where $\mathcal{L}(w, \alpha)$ is the task loss and $\textit{r}(\alpha)$ is an L1 regularization term used to restrict the total number of bits or computation cost within a given budget.
$\lambda_{r}$ is a scale factor.
Equation \ref{eq:bitsearch_reg} shows $\textit{r}(\alpha)$ when a computation cost constraint, i.e., a target number of bit operations $\textit{t\_bops}$ is given. 

\begin{equation}
\label{eq:bitsearch_reg}
    \textit{r}(\alpha) = \left| \sum_{i = 1}^{N}{op_i \cdot b^{w} \cdot \sum_{j = 1}^{B} \textit{hs}(\alpha_i)_j \cdot b_j } - \textit{t\_bops} \right|
\end{equation}
\vspace{+0.05cm}

where $op_i$ is the number of operations on layer $i$ and $b^{w}$ is the weight bit-width.\footnote{Note that we utilize a single bit-width of weight in the entire network. Refer to supplementary for weight mixed-precision.}
Since the per-layer bit-width of activation is finally determined by the bit-width of the branch having the top softmax score in Equation \ref{eq:bitsearch_darts}, it is required to obtain the bit-width of the top performer branch while making gradients flow through all the branches~\cite{gumbel} in order to learn architectural parameters $\alpha$.
To do this, we use straight-through-softmax trick on $\textit{hs}$ in Equation \ref{eq:bitsearch_reg}.
As Equation \ref{eq:bitsearch_hardsoftmax} shows, $\textit{hs}$ outputs one-hot vector of architectural parameter vector $\alpha$ (i.e., the element having the largest softmax function is assigned to one while the others to zero) in forward pass of training. 
However, in backward pass, we approximate the onehot to softmax results thereby allowing gradients to propagate through all the branches.

\begin{equation}
    \label{eq:bitsearch_hardsoftmax}
    \begin{gathered}
        \textit{hs}(\alpha_i) = \textit{onehot}(\alpha_i)  \\
        \nabla_{\alpha_j}\textit{hs}(\alpha_i) \approx \nabla_{\alpha_j}\textit{softmax}(\alpha_i) 
    \end{gathered}
\end{equation}
\vspace{+0.05cm}

The mixed-precision quantization of MetaMix looks similar to that of DNAS~\cite{dnas}.
Our key difference is the different utilization of branches to mitigate negative effect of bit-width on activation instability.
MetaMix forms a combination of single bit-width end-to-end networks (Figure \ref{fig:metamix_flow_diagram} left) to train the shared weights in bit-meta training.
Then the trained model, i.e., the meta-state model is fixed and only the architectural parameters of all bit-widths are trained in bit-search training (Figure \ref{fig:metamix_flow_diagram} right).
However, DNAS forms per-branch weights with assigned bit-width and architectural parameters which can maximize the activation instability on training weights.
After bit search, DNAS samples diverse architectures and exhaustively trains many model candidates, while MetaMix directly selects one final architecture for fine-tuning (in the weight training phase), which offers faster bit search.
As a result, MetaMix enables better mixed-precision model (71.94\% in 32.86GBOPs vs. 70.6\% in 35.17GBOPs on ResNet-18) and faster bit-width searching (GPU hours of 13.4h vs. 40h on ResNet-18) and re-training (fine-tuning in our case) compared to DNAS.

\section{Experiments}

\begin{table}[t]
    \begin{center}
        \begin{tabular}{|c|c|c|c|c|c|c|}
            \hline
            \multirow{2}{*}{Method} & \multirow{2}{*}{\makecell{Bits\\(A/W)}} & \multicolumn{2}{c|}{1st 8-bit} & \multicolumn{2}{c|}{last 8-bit} & \multirow{2}{*}{\makecell{Top-1\\(\%)}} \\
            \cline{3-6}
             & & A & W & A & W & \\
            \hline\hline
            Baseline & FP & & & & & 71.9 \\
            \hline
            PACT & 4/4 & & & & & 61.40 \\
            DSQ & 4/4 & & & & & 64.80 \\
            LLSQ & 4/4 & & & & & 67.37 \\
            LSQ & 4/4 & & & & & 69.5 \\
            LSQ+BR & 4/4 & & & & & 70.4 \\
            OOQ & 4/4 & \checkmark & \checkmark & \checkmark & \checkmark & 70.6 \\ 
            LCQ & 4/4 & & \checkmark &  & \checkmark & 70.8 \\
            SAT & 4/4 & & \checkmark & \checkmark & \checkmark & 70.8 \\
            APoT & 4/4 & & \checkmark &  & \checkmark & 71.0 \\
            PROFIT & 4/4 & & \checkmark & \checkmark & \checkmark & 71.56 \\
            PROFIT$^\dagger$ & 4/4 & \checkmark & \checkmark & \checkmark & \checkmark & 70.93 \\
            N2UQ & 4/4 &  &  &  &  & 72.1 \\
            N2UQ$^\dagger$ & 4/4 &  & \checkmark & \checkmark & \checkmark & 62.83 \\
            N2UQ$^\dagger$ & 4/4 & \checkmark & \checkmark & \checkmark & \checkmark & 52.12 \\
            \hline
            MetaMix & 3.98/4 & \checkmark & \checkmark & \checkmark & \checkmark & { \bf 72.60 } \\
            \hline
        \end{tabular}
    \end{center}
    \caption{ImageNet comparison with single-precision quantization results on MobileNet-v2 (PACT~\cite{pact}, DSQ~\cite{dsq}, LLSQ~\cite{llsq}, LSQ~\cite{lsq}, LSQ+BR~\cite{br}, OOQ~\cite{ooq}, LCQ~\cite{lcq}, SAT~\cite{sat}, APoT~\cite{apot}, PROFIT~\cite{profit}, N2UQ~\cite{n2uq}). In column `Bits', `A' is for activation bit-width and `W' is for weight bit-width. `1st 8-bit' and `last 8-bit' columns show whether the activation (A) or weight (W) of the 1st or last layer is quantized to 8-bit or not. PROFIT$^\dagger$ and N2UQ$^\dagger$ are re-implemented version.
    }
    \label{table:mbv2_fixed}
\end{table}

\begin{table}[t]
    \begin{center}
        \begin{tabular}{|c|c|c|c|c|c|c|c|c|}
            \hline
            \multirow{2}{*}{Method} & \multirow{2}{*}{\makecell{Bits\\(A/W)}} & \multicolumn{2}{c|}{1st 8-bit} & \multicolumn{2}{c|}{last 8-bit} & \multirow{2}{*}{\makecell{Top-1\\(\%)}} \\
            \cline{3-6}
             & & A & W & A & W & \\
            \hline\hline
            Baseline & FP & & & & & 75.3 \\
            \hline
            PACT & 4/4 & & & & & 70.16 \\
            DUQ & 4/4 & & \checkmark & \checkmark & \checkmark & 71.01 \\
            PROFIT & 4/4 & & \checkmark & \checkmark & \checkmark & 73.81 \\
            PROFIT$^\dagger$ & 4/4 & \checkmark & \checkmark & \checkmark & \checkmark & 71.57 \\
            \hline
            MetaMix & 3.83/4 & \checkmark & \checkmark & \checkmark & \checkmark & { \bf 74.24 } \\
            \hline
        \end{tabular}
    \end{center}
    \caption{ImageNet comparison with single-precision quantization results on MobileNet-v3 (large) (PACT~\cite{pact}, DUQ \& PROFIT~\cite{profit}).
    }
    \label{table:mbv3_fixed}
\end{table}

\begin{table}[t]
    \begin{center}
        \begin{tabular}{|c|c|c|c|c|c|c|}
            \hline
            \multirow{2}{*}{Method} & \multirow{2}{*}{\makecell{Bits\\(A/W)}} & \multicolumn{2}{c|}{1st 8-bit} & \multicolumn{2}{c|}{last 8-bit} & \multirow{2}{*}{\makecell{Top-1\\(\%)}} \\
            \cline{3-6}
             & & A & W & A & W & \\
            \hline\hline
            Baseline & FP & & & & & 70.5 \\
            \hline
            PACT & 4/4 & & & & & 69.2 \\
            LQ-Net & 4/4 & & & & & 69.3 \\
            DSQ & 4/4 & &  &  & & 69.56 \\
            LLSQ & 4/4 & & & & & 69.84 \\
            QIL & 4/4 & & & & & 70.1 \\
            APoT & 4/4 & & \checkmark &  & \checkmark & 70.7 \\
            LSQ & 4/4 & & & & & 71.1 \\
            LCQ & 4/4 & & \checkmark &  & \checkmark & 71.5 \\
            N2UQ$^\dagger$ & 4/4 &  &  &  &  & 71.91 \\
            N2UQ$^\dagger$ & 4/4 &  & \checkmark &  \checkmark & \checkmark & 70.60 \\
            N2UQ$^\dagger$ & 4/4 & \checkmark & \checkmark &  \checkmark & \checkmark & NC \\
            \hline
            MetaMix & 3.85/4 & \checkmark & \checkmark & \checkmark & \checkmark & { \bf 71.94 } \\
            MetaMix & 3.85/3 & \checkmark & \checkmark & \checkmark & \checkmark & { \bf 70.69 } \\
            MetaMix & 3.85/2 & \checkmark & \checkmark & \checkmark & \checkmark & { \bf 69.45 } \\
            \hline
        \end{tabular}
    \end{center}
    \caption{ImageNet comparison with single-precision quantization results on ResNet-18 (PACT~\cite{pact}, LQ-Net~\cite{lqnet}, DSQ~\cite{dsq}, LLSQ~\cite{llsq}, QIL~\cite{qil}, APoT~\cite{apot}, LSQ~\cite{lsq}, LCQ~\cite{lcq}, N2UQ~\cite{n2uq}).
    `NC' means not converged.
    }
    \label{table:res18_fixed}
\end{table}

\subsection{Training Details}
\label{section:training_details}

Table \ref{table:training_details} shows the details of training in our proposed method.
As the table shows, in the first epoch, we perform bit-meta training which learns full-precision weights and the step sizes of activations. Specifically, on each branch of bit-width (in Figure \ref{fig:searchstrategy_block}), the activation is quantized to its associated bit-width while its associated step size is being trained.
In the second epoch, we iterate bit-meta and bit-search training.
In bit-search training, we fix the full-precision weights and step sizes obtained in the bit-meta training and learn only the architectural parameters for per-layer bit-width probabilities.
After the per-layer bit-width is obtained, in the weight training phase, we fine-tune both network weights and the step sizes of weights and activations. 

We evaluate the proposed method on ImageNet-1K~\cite{imgnet}.
In MobileNet-v2 and v3, we use 8-bit, 4-bit, and 3-bit as candidate bit-widths of activations.
In ResNet-18, we use 8-bit, 4-bit, and 2-bit as candidates.
All weights are quantized to 4-bit.
We also quantize 1st and last layers to 8-bit for both weights and activations, which offers fully quantized networks.
Further details are in supplementary.

\subsection{Comparison on Single-Precision Quantization}

In this section, we compare single-precision quantization methods (PACT~\cite{pact}, LQ-Net~\cite{lqnet}, DSQ~\cite{dsq}, LLSQ~\cite{llsq}, QIL~\cite{qil}, LSQ~\cite{lsq}, LSQ+BR~\cite{br}, OOQ~\cite{ooq}, LCQ~\cite{lcq}, SAT~\cite{sat}, APoT~\cite{apot}, PROFIT \& DUQ~\cite{profit}, N2UQ~\cite{n2uq}) and ours using mixed-precision obtained with the target bit-width of 4-bit.\footnote{We use two decimal digits except some values having a single decimal digit in the original papers.}

Table \ref{table:mbv2_fixed} compares the accuracy of 4-bit level MobileNet-v2.
MetaMix (72.60\%) gives better accuracy than PROFIT (71.56\%) and N2UQ (72.1\%).
When the 1st layer input is quantized in PROFIT$^\dagger$, MetaMix outperforms it by a larger margin (72.60\% vs. 70.93\%).
N2UQ$^\dagger$ shows large accuracy drop with quantized inputs and weights of 1st and last layers.

Table \ref{table:mbv3_fixed} compares the accuracy of 4-bit level MobileNet-v3.
MetaMix outperforms PROFIT (74.24\% vs. 73.81\%) and, especially, PROFIT$^\dagger$ with 8-bit input by a large margin (74.24\% vs. 71.57\%).

Table \ref{table:res18_fixed} gives a comparison of 4-bit level ResNet-18.
The state-of-the-art method, N2UQ$^\dagger$, did not converged with quantized inputs and weights of 1st and last layers.
MetaMix (71.94\%) offers better accuracy than N2UQ$^\dagger$ (70.60\%) despite the fact that N2UQ$^\dagger$ adopts non-uniform quantization while not quantizing the input activation of 1st layer.

\subsection{Comparison on Mixed-Precision Quantization}

\begin{figure*}
    \begin{center}
        \includegraphics[width=\textwidth]{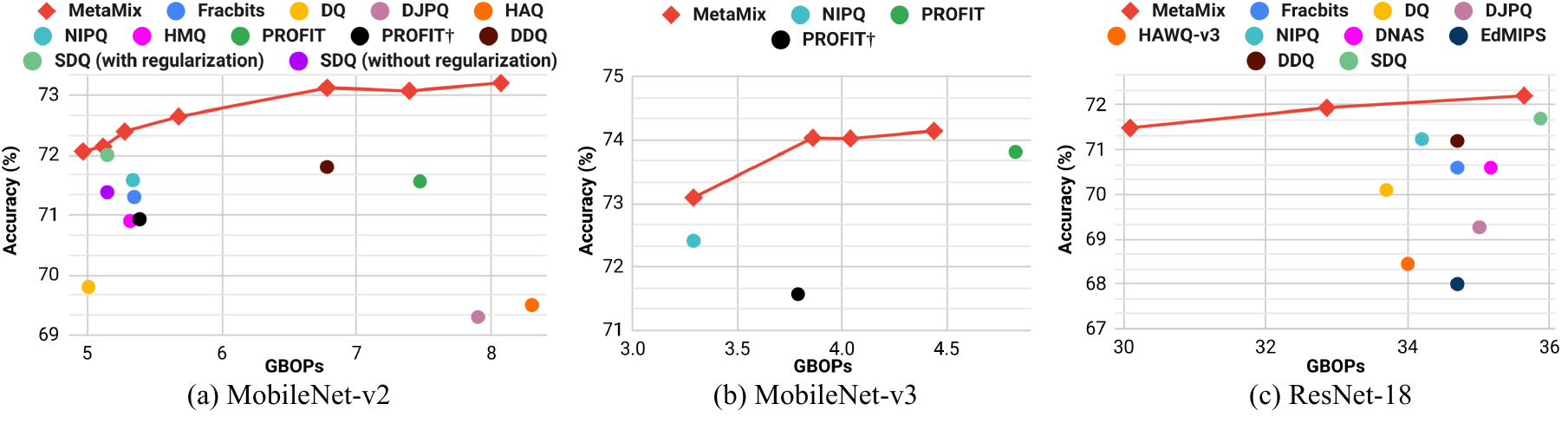}
    \end{center}
    \caption{
    ImageNet top-1 accuracy vs. BOPs on (a) MobileNet-v2, (b) MobileNet-v3 (large), (c) ResNet-18 (HMQ~\cite{hmq}, DQ~\cite{dq}, DJPQ~\cite{djpq}, HAQ~\cite{haq}, NIPQ~\cite{nipq}, Fracbits~\cite{fracbits}, DDQ~\cite{ddq}, SDQ~\cite{sdq}, DNAS~\cite{dnas}, HAWQ-v3~\cite{hawqv3}, EdMIPS~\cite{edmips} and state-of-the-art single-precision quantization PROFIT~\cite{profit}).
    PROFIT$^\dagger$ quantizes, in 8-bits, the input activation of 1st layer.
    }
    \label{fig:bops}
\end{figure*}

Figure \ref{fig:bops} shows the accuracy of mixed-precision quantization methods in diverse BOPs (bit operations) budget.
Note that MetaMix uses only a restricted set of bit-widths (8-bit, 4-bit, 3-bit in MobileNet-v2 and v3, 8-bit, 4-bit, 2-bit in ResNet-18) in search space while others use all the integer bits from 2-bit to 8-bit (HMQ, HAQ, Fracbits, DDQ, SDQ) or 2-bit to 10-bit (DQ, DJPQ, NIPQ).
The figure shows that MetaMix offers new state-of-the-art results while pushing the boundary of mixed-precision quantization.

In MobileNet-v2, MetaMix (72.14\% in 5.12GBOPs) shows comparable result to the state-of-the-art SDQ (72.0\% in 5.15GBOPs) and NIPQ (71.58\% in 5.34GBOPs) which uses more choices of integer bit.
The figure also shows a comparison with the single-precision state-of-the-art PROFIT.
MetaMix shows larger gain, i.e., 72.39\% in 5.28GBOPs (MetaMix) vs. 70.93\% in 5.39GBOPs (PROFIT$^\dagger$), with the quantized 1st layer input.

As reported in PROFIT~\cite{profit}, the depth-wise layers, near the input and output of the network, exhibit large activation instability due to weight quantization when they are quantized to 4-bit. PROFIT shows early stopping of those layers' training at the end of training improves final batch norm parameters.
We also tried to apply PROFIT to MetaMix and could not obtain improvements.
It is because those sensitive layers tend to be assigned high precision in mixed-precision quantization (as in Figure \ref{fig:bits_mbv2} (b)) thereby reducing their negative effect on activation instability.

In MobileNet-v3, MetaMix shows better accuracy (73.09\% in 3.29GBOPs) than mixed-precision NIPQ (72.41\% in 3.29GBOPs).
Compared to PROFIT (73.81\% in 4.83GBOPs) and PROFIT$^\dagger$ (71.57\% in 3.79GBOPs), MetaMix shows superior results (74.14\% in 4.44GBOPs and 73.09\% in 3.29GBOPs, respectively).

In ResNet-18, MetaMix pushes the boundary of mixed-precision quantization towards smaller BOPs.
Overall, MetaMix offers by more than 0.5\% better accuracy than other methods including SDQ (71.7\% in 35.87GBOPs) and NIPQ (71.24\% in 34.2GBOPs).

\subsubsection{Comparison with SOTA mixed-precision quantization}
SDQ applies strong regularization technique (i.e., 1.5\% improvement on ResNet-18).
In order to compare the sole effect of mixed-precision method itself, we compare MetaMix with SDQ without applying strong regularization.
In Figure \ref{fig:bops} (a), on MobileNet-v2, SDQ without strong regularization (71.38\% on 5.15G) shows dropped accuracy from SDQ with strong regularization (72.0\% on 5.15G).
However, MetaMix shows better accuracy (72.14\% on 5.12G) without any regularization.
Note that we also have far better accuracy in ResNet-18 (72.21\% in 35.64GBOPs vs. 71.7\% in 35.87GBOPs), even compared with strong regularization.

\subsection{Training Cost}
\label{section:training_cost}

\begin{table}[t]
    \begin{center}
        \begin{tabular}{|c|c|c|c|}
            \hline
            \multirow{2}{*}{Method} & \multicolumn{2}{c|}{Searching} & Re-training \\
            \cline{2-4}
            \multirow{2}{*}{} & GPU hours & Epochs & Epochs \\
            \hline \hline
            DNAS & 40 & 60 & 120 \\
            \hline
            SPOS & 312 & 120 & 240 \\
            \hline
            EdMIPS & 36 & 25 & 95 \\
            \hline
            MetaMix & 13.4 & 2 & 88 \\
            \hline
        \end{tabular}
    \end{center}
    \caption{Training cost comparison with NAS-based methods (DNAS~\cite{dnas}, SPOS~\cite{spos}, EdMIPS~\cite{edmips}) on ResNet-18 (Note that DNAS uses a small subset of ImageNet in searching). 
    }
    \label{table:nas_cost}
\end{table}

\begin{figure*}
    \begin{center}
    \includegraphics[width=0.9\linewidth]{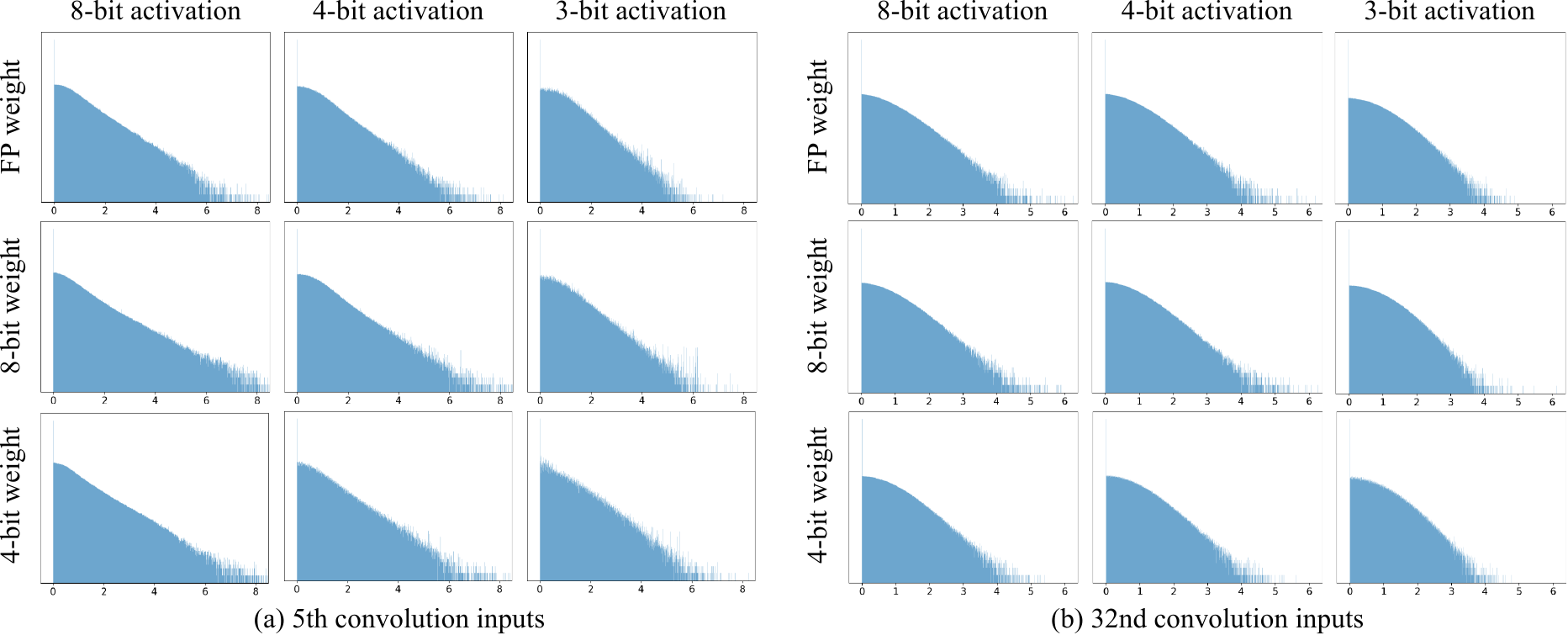}
    \end{center}
      \caption{
      Input activation distribution before quantizer with \{8, 4, 3\}-bit bit-meta training of MobileNet-v2.
      Formats are the same as in Figure \ref{fig:fixedbit_actdist}.
      All distributions show similar variances across different bit-widths of activations and weights in contrast to Figure \ref{fig:fixedbit_actdist}.
      This demonstrates that bit-meta training helps mitigate activation instability due to bit selection.
      }
    \label{fig:metamix_actdist}
\end{figure*}

\begin{figure}[t]
    \begin{center}
    \includegraphics[width=\linewidth]{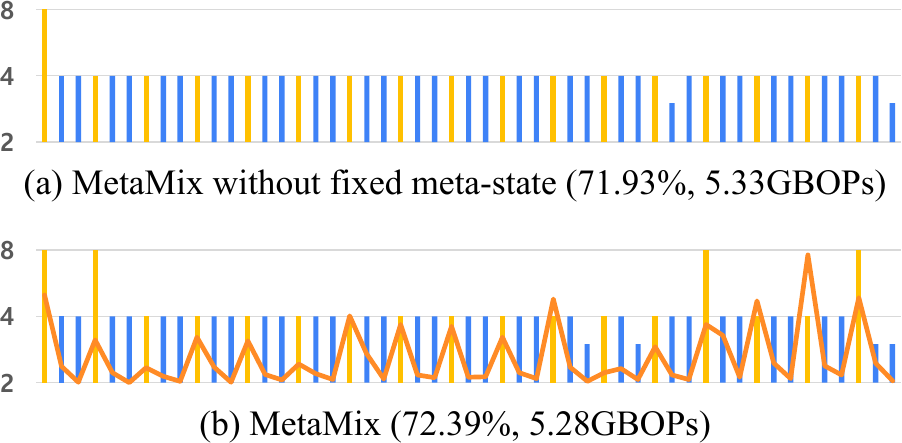}
    \end{center}
    \caption{
        Per-layer bit-width (yellow for depth-wise convolution layers and blue for the others) of MetaMix obtained (a) without and (b) with fixed meta-state on MobileNet-v2. (b) also shows the sum of hessian trace divided by the number of per-layer operations (orange line).
    }
    \label{fig:bits_mbv2}
\end{figure}

Table \ref{table:nas_cost} compares training cost with NAS based mixed-precision quantization methods.
The table shows MetaMix offers faster bit search ($>$3$\times$) and re-training (weight training in MetaMix) than existing methods.
The fast training advantages of MetaMix mostly come from the usage of meta-state in the bit selection phase.
Mixed-precision-aware initialization of weights in the weight training phase also contributes to faster training than the re-training of NAS-based methods which discard the weights obtained in bit search and re-train from scratch with the selected bit-width.

\section{Ablation Study}

\subsection{Effects of Meta-State Model}
\label{section:effects_meta-state}

Figure \ref{fig:metamix_actdist} shows the effects of bit-meta training on activations in MobileNet-v2.
The activation distributions exhibit much better consistency across different bit-widths than those in Figure \ref{fig:fixedbit_actdist}.
The consistency enables stable batch norm statistics when changing bit-width as in Figure \ref{fig:activation_stats} (bottom).
As such, the meta-state, which is the outcome of bit-meta training, offers consistent activation distributions across different activation bit-widths, which helps the subsequent bit-search training make high-quality bit selections by reducing the negative effect of activation instability due to bit selection.

As explained before, we use the fixed full-precision weights (i.e., meta-state) in bit-search training. 
Figure \ref{fig:metamix_actdist} also demonstrates the benefit of full-precision weights since, 
in the full-precision weight cases, activation distributions exhibit much stronger consistency across different activation bit-widths than in the cases of low bit-width weights.

\subsection{Effects of Fixed Meta-State Model}
\label{section:effects_fixed_meta_state}

In the second epoch of bit selection phase (in Table \ref{table:training_details}), we iterate bit-meta training and bit-search training.
The bit-search training utilizes the fixed full-precision weights after the previous bit-meta training. Thus, given the same mini-batch, there is no activation instability due to both bit selection and weight quantization in the subsequent bit-search training, which enables the bit-search training to benefit from the stable distribution of activation (as in the bottom of Figure \ref{fig:activation_stats} and Figure \ref{fig:metamix_actdist}) enabled by the bit-meta training.

Figure \ref{fig:bits_mbv2} illustrates the effects of fixed meta-state model in bit-search training on MobileNet-v2. The figure compares the per-layer bit-width results of two cases: (a) without and (b) with the fixed meta-state.
We set the same budget of computation cost (5.3GBOPs) in both cases.
When we do not fix but train the network weights (after initializing them with the meta-state) during bit-search training, as Figure \ref{fig:bits_mbv2} (a) shows, MetaMix tends to select 4-bit in most of layers.
However, as Figure \ref{fig:bits_mbv2} (b) shows, when the fixed meta-state is used, MetaMix tends to select higher bit-width for early and late depth-wise convolution layers which are known to be difficult to quantize in 4-bit while lowering the bit-widths of some intermediate layers, in return, finally meeting the budget.
As a result, MetaMix with the fixed meta-state in bit-search training gives better bit selections (72.39\%) than without the fixed meta-state (71.93\%).
Figure \ref{fig:bits_mbv2} (b) also shows the relationship between the selected per-layer bit-width, Hessian, and OPs. For details, refer to supplementary.

\section{Conclusion}

In this paper, we presented a novel training method, MetaMix, to address activation instability in mixed-precision quantization. It consists of bit selection and weight training phases.
In the bit selection phase, we determine the per-layer bit-width of activation by iterating the bit-meta training and the bit-search training, which reduce activation instability due to both bit selection and weight quantization thereby enabling fast and high-quality bit selection. The subsequent weight training phase offers fast fine-tuning of network weights and step sizes since they are initialized in a mixed-precision-aware manner in the previous bit selection phase. Our experiments on ImageNet show that MetaMix outperforms, in terms of accuracy vs. cost, single- and mixed-precision SOTA methods on efficient and hard-to-quantize models, i.e., MobileNet-v2 \& v3 and ResNet-18.

\section{Acknowledgments}
We appreciate valuable comments from Dr. Jiyang Kang, Wonpyo Park, and Yicheng Fan at Google.

\clearpage

\section{Supplementary Material}

\section{Activation Instability on Mixed-precision Quantization}

In this section, we provide more details about `Section: Activation Instability on Mixed-precision Quantization' of the main paper.
Activation instability due to bit selection results from the fact that the lower precision tends to incur the more variance in the quantized data. According to our observations, the increased activation variance of quantized layer also increases the scale factor of the previous batch norm layer as will be reported below.

\subsection{Larger variance of quantized value in lower precision}

\begin{figure*}[h]
    \begin{center}
        \includegraphics[width=0.99\linewidth]{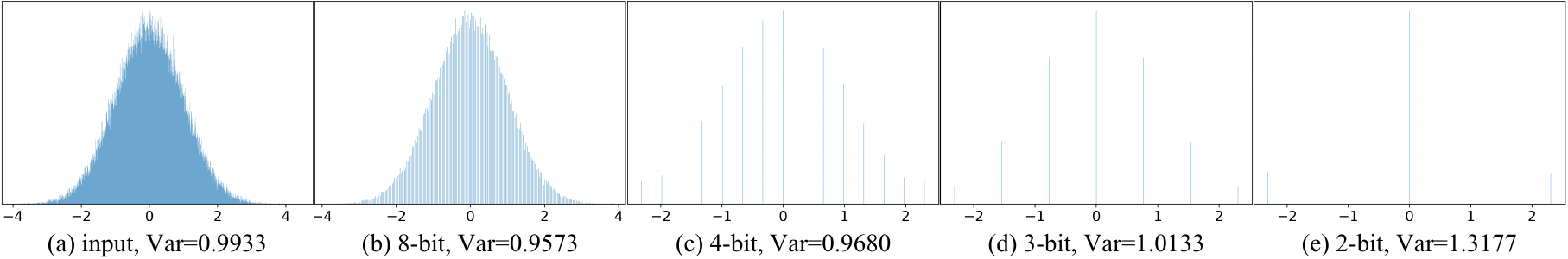}
    \end{center}
    \caption{
    A tensor sampled from Gaussian distribution and its variance (Var) (a) before quantization and after quantization to (b) 8-bit, (c) 4-bit, (d) 3-bit, and (e) 2-bit, respectively.
    X-axis is for values and y-axis is for frequency.
    }
    \label{fig:bit_variance}
\end{figure*}

Figure \ref{fig:bit_variance} exemplifies how the variance of activation distributions is affected by precisions.
We assume an input tensor sampled from Gaussian distribution (Figure \ref{fig:bit_variance} (a)) and quantize it to 8-bit, 4-bit, 3-bit and 2-bit (Figure \ref{fig:bit_variance} (b-e)), respectively.
As the figure shows, the quantization alters the variance of data.
Specifically, quantizing to the lower precision incurs the more variance in the quantized data.
Considering that the activation distributions tend to resemble Gaussian distribution, the figure demonstrates that the lower precision can incur the larger variance of activation in the quantized networks.

\section{Relationship between the selected per-layer bit-width, Hessian, and computation cost}

\begin{figure}
    \begin{center}
        \includegraphics[width=0.99\linewidth]{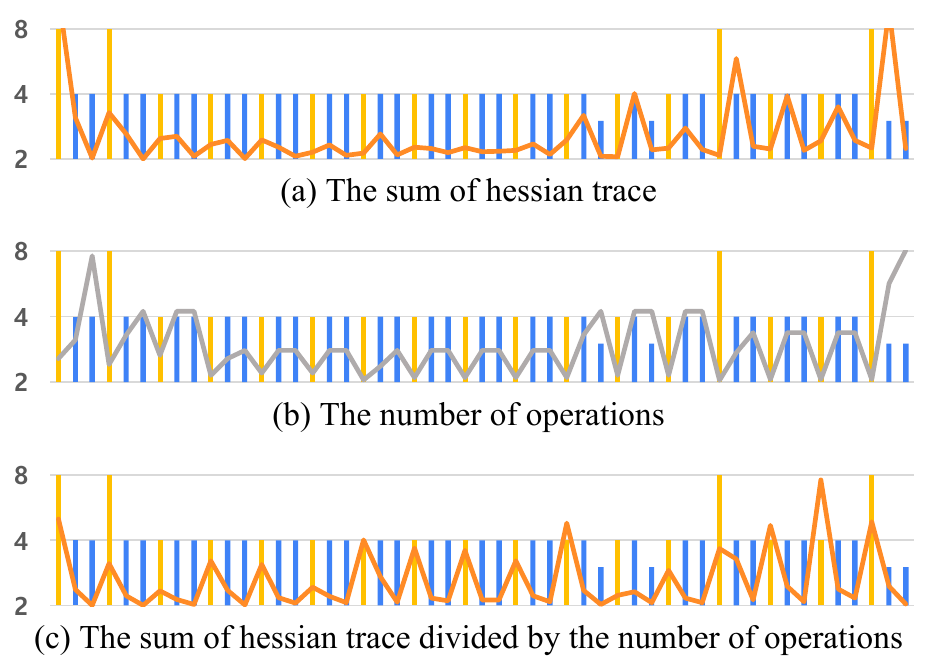}
    \end{center}
    \caption{
        Per-layer (a) sum of Hessian trace (orange line), (b) number of operations (gray line), and (c) the sum of Hessian trace divided by the number of operations (orange line) with per-layer bit-width (yellow for depth-wise convolution layers and blue for the others) of MetaMix on MobileNet-v2.
        In the main paper, we utilize (c) for sensitivity analysis with bit-width selections.
    }
    \label{fig:bits_hess_ops_mbv2}
\end{figure}

In this section, in addition to `Section: Effects of fixed meta-state model' of main paper, we provide more detailed analysis of bit selections in terms of Hessian and the number of operations (OPs).
Figure \ref{fig:bits_hess_ops_mbv2} shows the per-layer bit-widths under a constraint of computation cost in terms of OPs. 
Figure \ref{fig:bits_hess_ops_mbv2} (a) shows the sum of Hessian trace and the selected per-layer bit-widths in MobileNet-v2.
In the early and intermediate layers, both correlate with each other.
However, their correlation gets weaker in the late layers.
It is mainly because the late layers incur high computation cost as shown in 
Figure \ref{fig:bits_hess_ops_mbv2} (b).
As the two figures show, although the bit selection of MetaMix mostly follows Hessian metric, it selects low precision on the layers (mostly, the late layers) having high Hessian metric and high computation cost.

MetaMix aims at minimizing the training loss (Equation 3 in the main paper) which consists of task loss term and a regularization term (for computation cost in this case).
Both terms conflict against each other.
In order to visualize how MetaMix achieves the balance between the two,
Figure \ref{fig:bits_hess_ops_mbv2} (c) displays the sum of Hessian trace divided by the number of operations which follows the bit-width selection more faithfully than either Hessian metric or computation cost.
As such, MetaMix can offer per-layer bit-width selection result considering the sensitivity of layer in low precision and its computation cost.

\section{Extended Ablation Study}
\subsection{Effects of meta-state model}
\label{section:ablation_meta}

\begin{figure}[h]
    \begin{center}
    \includegraphics[width=1.0\linewidth]{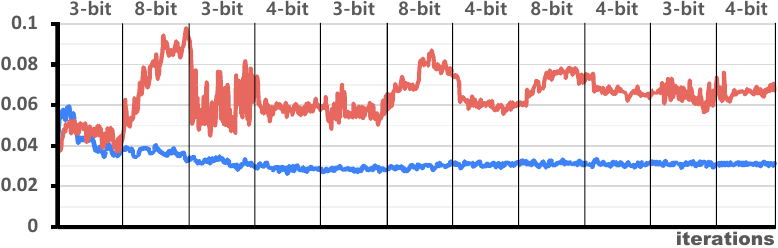}
    \end{center}
      \caption{
      KL divergence between FP and activation quantized models (input of 5th layer of MobileNet-v2) depending on whether bit-meta training is applied (blue) or not (red).
      }
    \label{fig:metamix_kldiv}
\end{figure}

In this section we present more ablation studies in addition to `Section: Effects of meta-state model' in main paper.
In Figure 1, 2 and 6 in main paper, the quantification of per-layer activation instability is limited to the activation distributions and batch norm statistics.
We provide KL divergence metric to further investigate the activation instability due to bit selection.
Figure \ref{fig:metamix_kldiv} shows KL divergence between FP and activation quantized models depending on whether bit-meta training is applied (blue) or not (red).
We randomly change the activation bit-width of the layer every epoch of training, same as Figure 2 in main paper.
Stable and low KL-div (blue) shows that bit-meta training enables robust activation distributions during bit-search.

\subsection{Effects of fixed meta-state model}
\label{section:ablation_fixedmeta}

In this section we present more ablation studies in addition to `Section: Effects of fixed meta-state model' in main paper.
We show further results of MobileNet-v3 (large).

\begin{figure}[h]
    \begin{center}
        \includegraphics[width=0.99\linewidth]{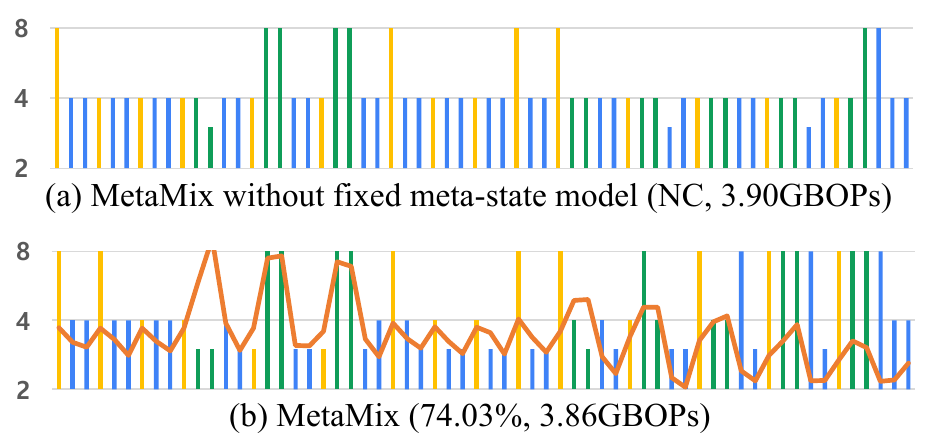}
    \end{center}
    \caption{
    Per-layer bit-width (yellow for depth-wise convolution layers, green for squeeze and excitation layers and blue for the others) of MetaMix obtained (a) without and (b) with fixed meta-state on MobileNet-v3.
    (b) also shows sensitivity metric (the sum of Hessian trace divided by the number of per-layer operations) in orange line.
    `NC' represents not converged.
    }
    \label{fig:bits_mbv3}
\end{figure}

Figure \ref{fig:bits_mbv3} illustrates the effects of fixed meta-state model in bit-search training on MobileNet-v3.
We set the same budget of computation cost (3.86GBOPs) in both cases.
The bit selection tendency is similar to that of MobileNet-v2.
By not fixing meta-state but training the network weights during bit-search training, as Figure \ref{fig:bits_mbv3} (a) shows, MetaMix tends to select 4-bit in most of layers.
However, as Figure \ref{fig:bits_mbv3} (b) shows, when the fixed meta-state is used, MetaMix tends to select higher bit-width for early and late depth-wise convolution layers and squeeze and excitation layers which are known to be difficult to quantize in 4-bit while lowering the bit-widths of some intermediate layers, in return, to meet the budget.
MetaMix with the fixed meta-state in bit-search training gives 74.03\% accuracy while the opposite case does not converge (denoted as NC in Figure \ref{fig:bits_mbv3} (a)).

\subsection{Effects of iterative bit-meta and bit-search training}

\begin{figure}[h]
    \begin{center}
        \includegraphics[width=0.99\linewidth]{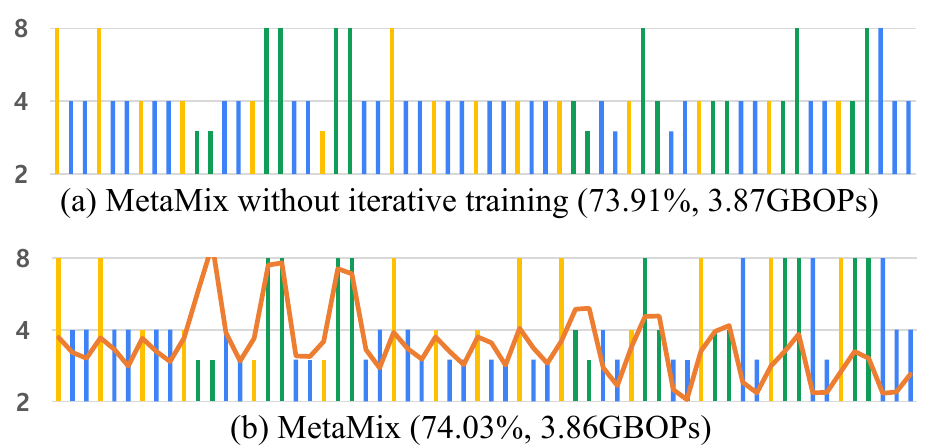}
    \end{center}
    \caption{
    Per-layer bit-width (yellow for depth-wise convolution layers, green for squeeze and excitation layers and blue for the others) of MetaMix obtained (a) without and (b) with iterative training on MobileNet-v3.
    (b) also shows sensitivity metric (the sum of Hessian trace divided by the number of per-layer operations) in orange line.
    }
    \label{fig:bits_mbv3_iterative}
\end{figure}

\begin{table*}[t]
    \begin{center}
        \begin{tabular}{|c|c|c|c|c|c|c|}
            \hline
            \multirow{2}{*}{Method} & \multicolumn{3}{c|}{Restricted - \{8-bit, 4-bit, 2-bit\}} & \multicolumn{3}{c|}{Original - \{2-bit to 8-bit\} or \{2-bit to 10-bit\} } \\
            \cline{2-7}
            \multirow{2}{*}{} & {Bits (A/W)}& {BOPs (G)} & {Top-1 (\%)} & {Bits (A/W)} & {BOPs (G)} & {Top-1 (\%)} \\
            \cline{2-7}
            \hline \hline
            DQ & $\cdot$ & 58.68 & 68.49 & $\cdot$ & 33.7 & 70.1\\
            \hline
            DJPQ & $\cdot$ & 35.45 & 69.12 & $\cdot$ & 35.01 & 69.27 \\
            \hline
            SDQ & 4/3.61 & $\cdot$ & 71.1 & 4/3.61 & 35.87 & 71.7 \\
            \hline
            MetaMix & 3.85/4 & 32.86 & {\bf 71.94} & $\cdot$ & $\cdot$ & $\cdot$ \\
            \hline
        \end{tabular}
    \end{center}
    \caption{ImageNet comparison with power-of-two-bit based mixed-precision quantization results on ResNet-18.
    In column `Bits’, `A’ is for activation bit-width and `W’ is for weight bit-width.
    }
    \label{table:pot_res_ori}
\end{table*}

MetaMix works with iterative bit-meta training and bit-search training in the 2nd epoch of bit selection phase.
Without iterative training, the training schedule can be conducted to apply bit-search training after the end of bit-meta training with totally fixed meta-state on the entire bit-search training.

Figure \ref{fig:bits_mbv3_iterative} displays the bit selection result and accuracy of MetaMix regarding iterative training on MobileNet-v3.
We set the computation cost budget as 3.86GBOPs in both with and without iterative training.
MetaMix tends to select more 8-bit selection in late layers with iterative training while lowering bit-widths of some intermediate layers.
Bit selection of early layers are similar in both with and without iterative training.
The iterative training proves effective (74.03\% vs 73.91\%), possibly due to the difference of bit selection in late layers.

\subsection{Power-of-two-bit based mixed-precision}

Hardware accelerators~\cite{nvturing,bitfusion} often support multiple power-of-two-bit precisions.
Thus, in order to utilize such hardware accelerators, the search space of bit-widths needs to be restricted to power-of-two bits, e.g., 8-bit, 4-bit, and 2-bit.
Table \ref{table:pot_res_ori} compares DQ~\cite{dq}, DJPQ~\cite{djpq}, SDQ~\cite{sdq} under power-of-two bits restriction (on column `Restricted') and their original results (on column `Original') with no restriction on ResNet-18.
Note that, as mentioned before, most of the methods report their main results by including all the integer bits from 2-bit to 8-bit~\cite{hmq,haq,fracbits,ddq,sdq} or 2-bit to 10-bit~\cite{dq,djpq,nipq}.
All the methods show degradation on power-of-two bits restriction compared to their original results with wider search space.
However, MetaMix, which originally has power-of-two bit-width search space, shows superior result even compared with original results of the others.

In case of MobileNet-v2 and v3, we used 3-bit instead of 2-bit in power-of-two bits restriction since 2-bit quantization shows inferior accuracy in our experiments.\footnote{Please note that MobileNets are well known to be hard to quantized in sub 4-bits.}
Considering the recent demands of efficiency, it is imperative to realize power-of-two-bit based mixed-precision quantization, covering down to 2-bit~\cite{basq}, on efficient networks like MobileNets, which is left as our future work.
\subsection{Extension to weight mixed-precision quantization}
We extend MetaMix to weight mixed-precision quantization by using bit-pairs, (A,W) with activation bit `A' and weight bit `W', as bit-width selection candidate and obtain the results as Table \ref{table:awmixed}.
We utilize 8-bit, 4-bit and 2-bit as candidate for 30.1GBOPs model and 6-bit, 4-bit and 2-bit as candidate for 31.0GBOPs model.
Note that the result of first row (71.49\% in 30.1GBOPs) is from Figure 5 (c) of main paper.
Extension to weight mixed-precision quantization gives far better results than activation only mixed-precision methods.

\begin{table}[h]
    \begin{center}
        \begin{tabular}{|c|c|c|}
            \hline
            (A,W) Search pairs & BOPs & Top-1 \\
            \hline\hline
            A=\{8,4,2\}, W=4 & 30.1G & 71.49\% \\
            \hline
            A=\{8,4,2\}, W=\{8,4,2\} $\Rightarrow$ 9-cases & 30.3G & 71.80\% \\ 
            \hline
            A=\{6,4,2\}, W=4 & 31.0G & 71.65\% \\
            \hline
            A=\{6,4,2\}, W=\{6,4,2\} $\Rightarrow$ 9-cases & 31.1G & 72.08\% \\
            \hline
        \end{tabular}
    \end{center}
    \caption{ImageNet comparison with MetaMix of both activation and weight mixed-precision quantization results on ResNet-18.}
    \label{table:awmixed}
\end{table}

\section{Training in detail}
In this section we show further details in `Section: Training details' in main paper.
We adopt LSQ~\cite{lsq} quantizer for all input activations and weights.
For all experiments, we use a batch size of 256 with an initial learning rate of 0.0005, Adam optimizer~\cite{adam} with cosine learning rate decay without restart, and knowledge distillation~\cite{kd} with ResNet-101 as teacher.

As shown in Table 1 of main paper, we utilize different training epochs for different networks.
For MobileNet-v2 and v3, we use 2-epochs to bit-selection phase and 138-epochs to weight training phase so that total epochs are 140-epochs.
For ResNet-18, we use 2-epochs to bit-selection phase and 88-epochs to weight training phase so that total epochs are 90-epochs.

Table \ref{table:impl_detail} shows $\lambda_r$, scale factor for L1 regularization, for different target networks.

\begin{table}[h]
    \begin{center}
        \begin{tabular}{|c|c|}
            \hline
            Network & $\lambda_r$ \\
            \hline \hline
            MobileNet-v2 & 1 \\
            \hline
            MobileNet-v3 & 4 \\
            \hline
            ResNet-18 & 0.4 \\
            \hline
        \end{tabular}
    \end{center}
    \caption{Scale factor for L1 regularization term on MobileNet-v2, MobileNet-v3 (large) and ResNet-18.}
    \label{table:impl_detail}
\end{table}

\section{Results in detail}
In addition to `Section: Comparison on mixed-precision quantization' (Figure 5) in the main paper, Table \ref{table:bops} shows top-1 accuracy and GBOPs in Figure 5 of the main paper.

\begin{table}[h]
    \begin{center}
        \begin{tabular}{|c|c|c|}
            \hline
            Network & BOPs (G) & Top-1 (\%) \\
            \hline \hline
            \multirow{7}{*}{MobileNet-v2} & 8.07 & 73.20 \\
            \multirow{7}{*}{} & 7.39 & 73.07 \\
            \multirow{7}{*}{} & 6.78 & 73.12 \\
            \multirow{7}{*}{} & 5.68 & 72.64 \\
            \multirow{7}{*}{} & 5.28 & 72.39 \\
            \multirow{7}{*}{} & 5.12 & 72.14 \\
            \multirow{7}{*}{} & 4.97 & 72.06 \\
            \hline
            \multirow{4}{*}{MobileNet-v3} & 4.44 & 74.14 \\
            \multirow{4}{*}{} & 4.04 & 74.02 \\
            \multirow{4}{*}{} & 3.86 & 74.03 \\
            \multirow{4}{*}{} & 3.29 & 73.09 \\
            \hline
            \multirow{3}{*}{ResNet-18} & 35.64 & 72.21 \\
            \multirow{3}{*}{} & 32.86 & 71.94 \\
            \multirow{3}{*}{} & 30.09 & 71.49 \\
            \hline
        \end{tabular}
    \end{center}
    \caption{ImageNet top-1 accuracy vs. BOPs on MobileNet-v2, MobileNet-v3 (large) and ResNet-18.}
    \label{table:bops}
\end{table}

\section{Comparison in detail}

\begin{table*}[t]
    \begin{center}
        \begin{tabular}{|c|c|c|c|c|c|c|c|c|c|}
            \hline
            \multirow{2}{*}{Method} & \multirow{2}{*}{\makecell{Bits\\(A/W)}} & \multicolumn{2}{c|}{1st 8-bit} & \multicolumn{2}{c|}{last 8-bit} & \multicolumn{2}{c|}{Top-1 (\%)} \\
            \cline{3-8}
             & & A & W & A & W & Main paper (N2UQ) & Reproduction (N2UQ$^\dagger$) \\
            \hline\hline
            Baseline & FP & & & & & 71.8$^\star$ & 70.5 \\
            \hline 
            N2UQ / N2UQ$^\dagger$ & 4/4 & & & & & 72.9$^\star$ & 71.91 \\ 
            N2UQ / N2UQ$^\dagger$ & 4/4 & & \checkmark & \checkmark & \checkmark & 70.83 & 70.60 \\ 
            N2UQ / N2UQ$^\dagger$ & 4/4 & \checkmark & \checkmark & \checkmark & \checkmark & 70.27 & NC \\
            \hline
            MetaMix & 3.85/4 & \checkmark & \checkmark & \checkmark & \checkmark & \multicolumn{2}{c|}{{ \bf 71.94 }} \\
            \hline
        \end{tabular}
    \end{center}
    \caption{
    ImageNet top-1 accuracy of N2UQ with pre-trained FP model used in original implementation (column `Main paper') and reproduction (column `Reproduction') on ResNet-18.
    In column `Bits’, `A’ is for activation bit-width and `W’ is for weight bit-width. 
    `1st 8-bit’ and `last 8-bit’ columns show whether the activation (A) or weight (W) of the 1st or last layer is quantized to 8-bit or not.
    `NC' means not converged.
    ($^\star$ is the number from N2UQ main paper~\cite{n2uq}).
    }
    \label{table:n2uq_reprod}
\end{table*}

\subsection{Re-implementations}
\subsubsection{PROFIT~\cite{profit}}
We utilize official code\footnote{https://github.com/EunhyeokPark/PROFIT} and further quantize inputs of 1st layer.
\subsubsection{N2UQ~\cite{n2uq}}
We utilize official code\footnote{https://github.com/liuzechun/Nonuniform-to-Uniform-Quantization} and further quantize inputs and weights of 1st and last layer.
Furthermore, in ResNet-18, since N2UQ utilizes much higher pre-trained FP model (71.8\%) compared to others and MetaMix (70.5\%), we retrain with same pre-trained FP model as others and MetaMix for fair comparison.
Table \ref{table:n2uq_reprod} compares N2UQ results with the pre-trained FP model used in original implementation (N2UQ, column `Main paper' with 71.8\% baseline) and the pre-trained FP model used in our reproduction (N2UQ$^\dagger$, column `Reproduction' with 70.5\% baseline).
The results of N2UQ are as higher as the improved baseline (also making not converged model converge) compared to N2UQ$^\dagger$.
However, when we quantize inputs and weights of 1st and last layer of N2UQ, it shows large accuracy degradation compared to not quantizing them (70.27\% vs 72.9\%).
Although N2UQ starts from better baseline, when the inputs and weights of 1st and last layer are quantized, MetaMix outperforms it by a larger margin (71.94\% vs 70.27\%).

\subsection{BOPs calculation}
We calculate computation cost, BOPs (bit operations), in the same manner as existing mixed-precision works~\cite{dnas,spos,fracbits}.
Given per-layer activation bit-width $b_i$, Equation \ref{eq:bitops} shows the BOPs calculation.

\begin{equation}
\label{eq:bitops}
    \mbox{BOPs} = \sum_{i = 1}^{N}{ op_i \cdot b^w \cdot b_i }
\end{equation}

where $b^w$ is a single bit-width of weight in the entire network and $op_i$ is the number of operations on layer $i$.
Note that we utilize a single bit-width of weight in the entire network.

\clearpage

\bibliography{aaai24.bbl}

\end{document}